\documentclass[10pt,twocolumn,letterpaper]{article}

\usepackage{cvpr}              

\usepackage{graphicx}
\usepackage{amsmath}
\usepackage{amssymb}
\usepackage{booktabs}
\usepackage[accsupp]{axessibility} 

\usepackage[pagebackref,breaklinks,colorlinks]{hyperref}

\usepackage{algorithm}
\usepackage{algorithmic}

\usepackage{amsmath}
\usepackage{amssymb}

\usepackage{color}
\usepackage{subfigure}
\usepackage[switch]{lineno}
\newcommand*{\affaddr}[1]{#1} 
\newcommand*{\affmark}[1][*]{\textsuperscript{#1}}

\newcommand\blfootnote[1]{%
\begingroup
\renewcommand\thefootnote{}\footnote{#1}%
\addtocounter{footnote}{-1}%
\endgroup
}

\usepackage[capitalize]{cleveref}
\crefname{section}{Sec.}{Secs.}
\Crefname{section}{Section}{Sections}
\Crefname{table}{Table}{Tables}
\crefname{table}{Tab.}{Tabs.}

\def\cvprPaperID{4045} 
\def\confName{CVPR}
\def\confYear{2022}

\begin{document}

\title{MogFace: Towards a Deeper Appreciation on Face Detection }


\author{
Yang Liu \affmark[1] \quad 
Fei Wang \affmark[1] \quad
Jiankang Deng \affmark[2] \quad
Zhipeng Zhou \affmark[1] \quad
Baigui Sun \affmark[1] \quad
Hao Li \textsuperscript{*} \affmark[1] \quad \\
\affaddr{\affmark[1]Alibaba Group} \qquad
\affaddr{\affmark[2]Imperial College London} \qquad \\
}

\maketitle

\begin{abstract}
Benefiting from the pioneering design of generic object detectors, significant achievements have been made in the field of face detection.
Typically, the architectures of the backbone, feature pyramid layer, and detection head module within the face detector all assimilate the excellent experience from general object detectors. 
However, several effective methods,  including label assignment and scale-level data augmentation strategy, fail to maintain consistent superiority when applying on the face detector directly. 
Concretely, the former strategy involves a vast body of hyper-parameters  and the latter one suffers from the challenge of scale distribution bias between different detection tasks, which both limit their generalization abilities.
Furthermore, in order to provide accurate face bounding boxes for facial down-stream tasks, the face detector imperatively requires the elimination of false alarms.
As a result, practical solutions on label assignment, scale-level data augmentation, and reducing false alarms are necessary for advancing face detectors. 
In this paper, we focus on resolving three aforementioned challenges that exiting methods are difficult to finish off and present a novel face detector, termed MogFace. In our Mogface, three key components, Adaptive Online Incremental Anchor Mining Strategy, Selective Scale Enhancement Strategy and Hierarchical Context-Aware Module, are separately proposed to boost the performance of face detectors.
Finally, 
to the best of our knowledge, our MogFace is the best face detector on the Wider Face leader-board, achieving all champions across different testing scenarios. The code is available at \url{https://github.com/damo-cv/MogFace}.
\end{abstract} 

\blfootnote{\textsuperscript{*} Corresponding Author. \\
\qquad Email:  ly261666@alibaba-inc.com\\
} 
\section{Introduction}

\label{sec:introduction}
Face detector, predicting location coordinates of face boxes, serves as the fundamental step for many facial downstream tasks, including face alignment \cite{bulat2017far}, face recognition \cite{deng1801arcface} \cite{wang2021efficient} and face attribute analysis \cite{shu2021learning}. In the past few years, we have witnessed the quick development on the general object detectors, deriving from the Fast-RCNN \cite{girshick2015fast} and SSD \cite{liu2016ssd} to Retinanet \cite{lin2017focal} and DERT \cite{carion2020end}. 
Motivated by this, 
 state-of-the-art face detectors adopt the great architecture designs from general object detectors, such as Feature Pyramid Network \cite{lin2017feature} and One-stage Single-Shot framework \cite{lin2017focal}. 

\begin{figure*}[t]
    \subfigure[]{
    \includegraphics[height=0.13\textwidth]{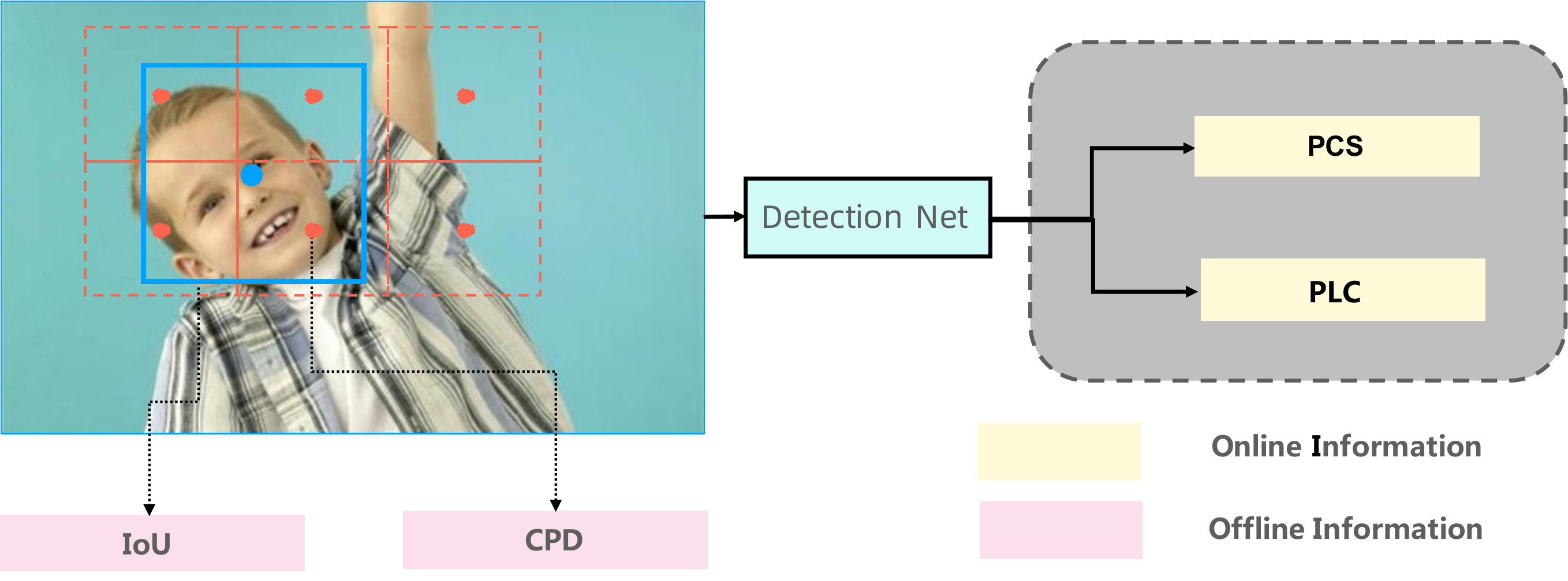}
    \label{img1_a}
    }
    \subfigure[]{
    \includegraphics[height=0.13\textwidth]{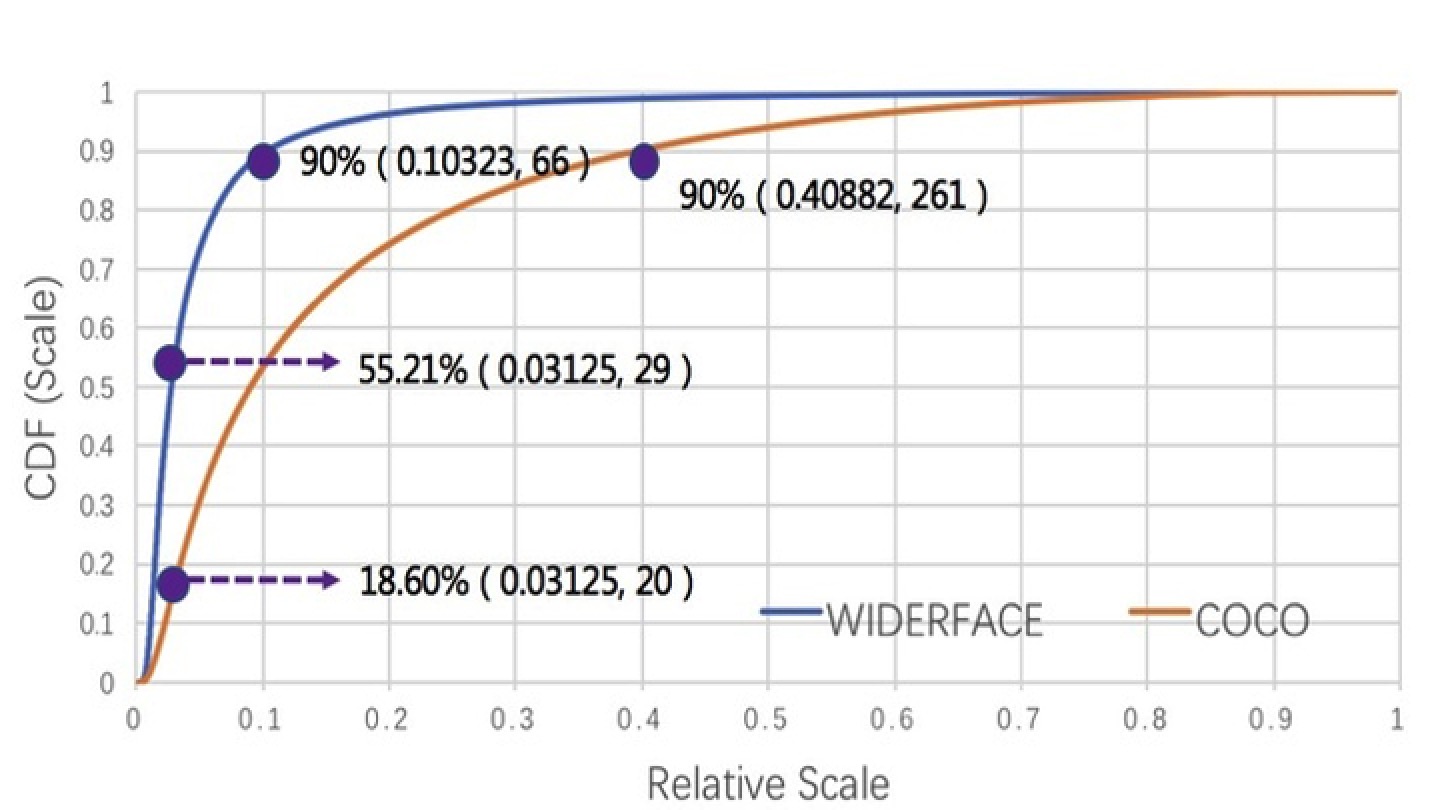}
    \label{img1_b}
    }
    \subfigure[]{
    \includegraphics[height=0.13\textwidth]{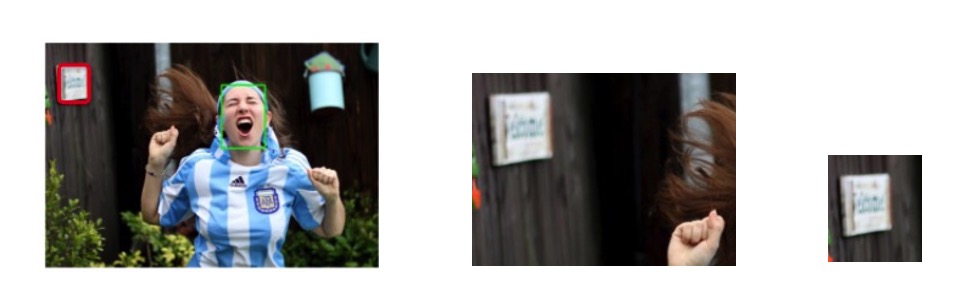}
    \label{img1_c}
    }
    \caption{Motivation illustration. (a) 
    Online and offline information both can be adopted as criterion to determine the boundary between positive and negative anchors. But how to effectively and adaptively combine them remains a huge challenge.
    (b) Cumulative density curve of face or object scale relative to the fixed scale (640). In the Wider face and COCO dataset, almost 55\% and 18\% ground-truth scale is less than 20, demonstrating that compared to generic object detector, a more severe scale variance challenge is occurred on the task of face detection. (c) For the same detector, we discover the top-left calendar is a false alarm in the left image, while  the top-left calendar in the other two images are not.}
    \label{img1}
\end{figure*}

However, label assignment and scale-level data augmentation strategy \footnote{enriches the scale distribution of the training data to resolve scale variance challenge.}, achieving great superiority on the task of generic object detection, bring rare gains on face detectors . On the one hand , the designation of former strategy involves a vast body of hyper-parameters (e.g. $K$ in ATSS \cite{zhang2020bridging}, $\alpha$ in OTA \cite{ge2021ota} ) , which limits its generalization ability.
 On the other hand, as shown in Fig. ~\ref{img1_b}, compared with generic object detector, face detector confronts more severe scale variance challenge. Uniform sampling based scale-level data augmentation strategies (e.g. multi-scale training  and random square crop \cite{liu2016ssd}), serving as the main scale enhancement methods on generic object detectors \cite{ren2015faster, lin2017focal, zhang2020bridging},  fail to provide effective scale information for face detector (more analysis can be seen in the supplementary material).
 Furthermore, the face detector is a real-world application that emphasizes reducing the number of false alarms urgently. Therefore, how to distinguish false alarms away from true positive faces is an another distinctive challenge on the task of face detection.
 

Based on the above analysis, we discover that label assignment strategy, scale-level data augmentation strategy and eliminating false alarms have a huge potential for constructing a high-performance face detector. 
Then we perform a systematically quantitative and qualitative analysis  on 3 aforementioned perspectives to provide some intrinsic insights.

\noindent\textbf{Label Assignment.} Label assignment strategies adopt predefined rules to match ground-truth (gt) or background for each anchor. As shown in the Fig. ~\ref{img1_a}, the designation of predefined rules highly depends on  offline and online information. Offline information contains Intersection-over-Union (IoU) and Center Point Distance (CPD) between gt and anchor, which can be computed during the process of data preparation. Online information consists of the predicted classification scores (PCS) and the predicted location coordinates (PLC), which can be extracted at the end of forward propagation. Traditional label assignment strategies adopt offline information as threshold criterion for $pos/neg$ anchors division, e.g. IoU in retinanet \cite{lin2017focal}, faster-rcnn \cite{ren2015faster}, IoU and CPD in ATSS \cite{zhang2020bridging}. Recently, Hambox \cite{liu2019hambox} further points out the effectiveness of online information and put forwards an online high-quality anchor mining strategy to utilize the PLC. OTA \cite{ge2021ota} formulates the assigning procedure as an optimal transport problem, where the cost function is  designed by the weighed combination of CPD, PCS and PLC. 

However, there exist two drawbacks lying behind current label assignment strategies: 1) Online information cannot provide high-confidence matching information as well as offline one. Thus, it will result in the emergence of sub-optimal label assignment strategy when encouraging online rules to serve as main metric on distinguishing positive and negative anchors like OTA and Hambox.
2) The selection of hyper-parameter in the recent label assignment strategies frequently goes through constant trials and errors, making it difficult for transferring different detection tasks, e.g. from general object detection to face detection.
In this paper, we address the aforementioned issues by proposing an adaptive online incremental anchor mining strategy (Ali-AMS), which is based on the standard anchor matching strategy adopted in retinanet \cite{lin2017focal} and further compensates outlier ground-truths with incremental anchors at the end of forward propagation. In our Ali-AMS, two key components, quality assessment based anchor mining strategy and pyramid-level consistency principle, are proposed to mine and assign the high-quality anchor adaptively. Concretely, 
the former strategy regards the PCS as the quality assessment criterion to re-sort the anchor mined with the CPD and IoU information; then, the latter principle guarantees that ground-truths located at the same pyramid layer can match the same number of anchors. The motivation and more details of our Ali-AMS can be seen in section \ref{sec:Ali-AMS}.

\noindent\textbf{Scale-level Data Augmentation. }
Generic object detector frequently introduces scale-level data augmentation strategy  to resolve extreme scale variance on the COCO dataset.  However, the most authoritative face detection dataset Wider Face \cite{yang2016wider} contains more severe scale variance than COCO \cite{lin2014microsoft}. As shown in Fig. ~\ref{img1_b}, we display the scale distribution of ground-truths on the COCO and Wider Face benchmark, respectively. Comparing with COCO dataset which is famous with extreme scale variance, Wider Face has more rigorous scale distribution, where contains almost 55\% small scale faces \footnote{the scale of ground-truth is less than 20}. 

To resolve extreme scale variance challenge, there exist 3 widely-adopted data augmentation strategies, including Multi-scale-training (MST), Random Square Crop (RSP) and Data-anchor-sampling (DAS).  Multi-scale-training strategy, resizing each image into a random scale selected from fixed scale range, frequently serves as the optimal solution on handling with scale variance problem, which has demonstrated its significance in many technology reports on the COCO detection challenge. Random Square Crop strategy, cropping the square area from a given image randomly, is a main-stream scale-level data augmentation strategy on the task of face detection \cite{deng2019retinaface}, \cite{zhang2017s3fd}. Data-anchor-sampling strategy \cite{tang2018pyramidbox} aims to introduce more small scale faces by resizing each image into a smaller scale.

MST and RSP are both designed from uniform sampling perspective while DAS focuses on generating many small faces. Meanwhile, in the Wider Face training set, almost 55\% face scale is less than 20, making uniform sampling based augmentation strategies generate a large number of small faces. As a result, MST and RSP both have great detection ability on small faces. 
This raises a worth solving problem: How to increase the detection ability on the middle and large scale faces since the training set only contains a small proportion large faces (10\%) ?  
In this paper, we investigate this question by analyzing the relationship between the performance of each pyramid layer and the number of ground-truths it matches. 
Based on comprehensive quantitative and qualitative analysis in  supplementary material, we unexpected discover that it
 is not  the more ground-truths that is matched in a single pyramid layer, the greater performance of this pyramid layer.
 As a result, this phenomenon releases an amazing conclusion that in order to improve the representation of certain pyramid layer, the number of ground-truths matched in this layer should be appropriate instead of 'the more, the better'. Under the guidance of this meaningful conclusion, we propose a simplex selective scale enhancement strategy for detecting large scale face accurately on the basis of prior statistical result, which controls the ground-truths distribution  to improve the deep pyramid layer representation, achieving the best detection performance on large and middle scale faces. To the best of our knowledge, our selective scale enhancement strategy is the first novel work to consider the relationship between the performance of each pyramid layer and the number of ground-truths it matches, which provides a solid knowledge on how to mine the learning capacity of certain pyramid layer.

\noindent\textbf{Eliminating False Alarms.}
Reducing the number of false alarms is vitally important for the real-world face detector. The common solution is to introduce additional training data with false alarms, which helps the detector acquire more knowledge on the property of false alarms. However, collecting  extra training data is labor-intensive such that the solution without extra data is deserved exploring.  
In this paper, we  present a  Hierarchical Context-Aware Module (HCAM) to help false alarms away from ground-truths, which  explicitly encodes neighbour context information into high-confidence anchors \footnote{high-confidence anchors contains correct predicted positive anchors and false predicted negative anchors}. 
The effectiveness of neighbour context information can be seen in Fig. \ref{img1_c}, we send the left image into the Hambox \cite{liu2019hambox} face detector and find a top-left false alarm. However, when we crop this false alarm with expanding context area (middle image) and less expanding context area (right image), we unexpectedly discover that the same detector believes there is no false positives in these two images. Moreover, we utilize Hambox face detector to find all false alarms in the Wider Face validation dataset, almost 95\% false alarms are disappeared when adopting similar operations like above. This phenomenon demonstrates that the appropriate neighbour context information is conducive to eliminating false alarms. 

In summary, our contribution can be summarized as:
\begin{itemize}
\setlength{\itemsep}{0pt}
    \item Presenting 3 worthy of in-depth research topics on the task of face detection, including Label Assignment, Scale-level Data Augmentation and Eliminating False Alarms.
    \item Proposing Adaptive Online Incremental Anchor Mining Strategy, Selective Scale Enhancement Strategy, Hierarchical Context-Aware module respectively to construct a promising face detector, termed as MogFace. 
    \item Achieving state-of-the-art results in all popular face detection  benchmarks, including  Wider Face, AFW, FDDB and Pascal Face. 
\end{itemize}


\section{Related Work}
Recently, a large number of face detectors \cite{liu2020hambox, liu2020bfbox, guo2021sample, tang2018pyramidbox, deng2019retinaface, li2019dsfd} has been proposed to advance face detection community. In this section, we mainly review the related work from two following perspectives, label assignment and scale-level data augmentation strategies.

\noindent\textbf{Label Assignment.} In the field of face detection, zhang et.al. \cite{zhang2017s3fd} propose a scale compensation anchor matching strategy to increase the matched anchors of outer faces by reducing the IoU threshold. Liu et al. \cite{liu2019hambox} discover  that  some  negative  anchors  have  stronger  localization  ability  than  positive  ones  and  further  propose an “online high-quality anchor mining strategy” by compensating outer faces with high-quality negative anchors. In the field of object detection, ATSS \cite{zhang2020bridging} automatically selects
positive and negative samples according to statistical
characteristics of object. OTA \cite{ge2021ota} formulates the label assigning procedure as an optimal transport problem, which converts the best assignment solution into solving the optimal transport plan at minimal transportation costs. 
 
\noindent\textbf{Scale-level Data Augmentation. } SNIP \cite{singh2018analysis} introduces a Scale Normalization for Image Pyramids (SNIP) strategy which selectively back-propagates the gradients of object instances of different sizes as a function of the image scale. Li et.al \cite{li2019scale} proposes a novel Trident Network to generate scale-specific feature maps with a uniform representational power. Data-anchor-sampling \cite{tang2018pyramidbox} augments the training samples to increase the diversity of training data for smaller faces. Zhang et.al \cite{zhang2017s3fd} proposes a scale-equitable face detection framework to handle different scales of faces. ASFD \cite{zhang2020asfd} introduces an automatic feature enhance module to allow  multi-scale feature fusion efficiently. 


\section{Method}
\label{sec:Method}
In this section, we consecutively introduce three components of our MogFace, including Adaptive Online Incremental Anchor Mining Strategy, Selective Scale Enhancement Strategy and Hierarchical Context-Aware Module.

\subsection{Adaptive Online Incremental Anchor Mining Strategy}
\label{sec:Ali-AMS}
Algorithm \ref{alg:ali_ams} describes how our Ali-AMS compensates high-quality anchors for outer ground-truths, when given an input image. Concretely, our Ali-AMS consists of pyramid-level consistency principle and quality assessment based anchor mining strategy, which identifies the number of anchors matched with each ground-truth among all pyramid layers and mines the high-quality anchors for outer ground-truth, respectively. The first principle (line 2-3) is described as follows: on each pyramid level, we first find the set of ground-truths ($\mathcal{G}$) matched in this pyramid layer based on the standard anchor matching strategy \cite{lin2017focal} and compute the maximum number ($\mathcal{T}$) of anchors that matched in $\mathcal{G}$. For each gt ($g$) in $\mathcal{G}$, if the number of anchors matched with $g$ is less than $\mathcal{T}$, the number of anchors that the $g$ matches will be  compensated to $T$ with following metric.  The description of  second strategy is from line 8 to 12: For each ground-truth that needs to be matched with incremental anchors, the select process of compensated anchors contains three following steps: 1) select $\mathcal{T}$ anchors from the view of CPD and IoU separately. 2) Sort all 2 * $\mathcal{T}$ anchors by the predicted classification score 3) Compensate top-$\mathcal{N}_{g}$ anchors for the outer ground-truth. Note that the number of top-$\mathcal{N}_{g}$ is computed in the first principle adaptively. The motivation behind our Ali-AMS is explained as follows.

\begin{algorithm}[t!]
\small
\caption{Adaptive Online Incremental Anchor Mining Strategy (Ali-AMS)} 
\label{alg:ali_ams} 
\begin{algorithmic}[1]
\REQUIRE ~~\\
$\mathcal{A}$ is a dict, key is ground-truth, value is the number of anchors matched with this ground-truth with standard anchor matching strategy. \\
\vspace{2mm}
\FOR{pyramid layer $pi$ in $[p2, p3, p4, p5, p6, p7]$}
\STATE $\mathcal{G} \leftarrow{}$ ground-truths that matched in the $pi$ \\
\STATE $\mathcal{T} \leftarrow{}$ maximum number of anchors that matched in $\mathcal{G}$ with standard anchor matching strategy. \\
\FOR{each ground-truth $g \in \mathcal{G}$} 
\IF{$\mathcal{A}[g] == \mathcal{T}$}
\STATE continue
\ENDIF \\
\STATE compute the number of compensated anchors for $g$: $\mathcal{N}_g = \mathcal{T} - \mathcal{A}[g]$;  \\
\STATE $cpd \leftarrow{}$ select $\mathcal{T}$ anchors whose centers are closest to the center of ground-truth $g$ based on L2 distance; \\
\STATE $iou \leftarrow{}$ select $\mathcal{T}$ anchors whose have the highest iou with ground-truth; \\
\STATE $conf\_candidate \leftarrow{}$ sort the anchor in $iou$ and $cpd$ according to predicted classification scores; \\
\STATE  select top-$\mathcal{N}_g$ confident anchors from $conf\_condidate$ to serve as positive anchors for $g$; \\ 
\ENDFOR
\ENDFOR 

\end{algorithmic}
\end{algorithm}
\noindent\textbf{Adopting predicted classification score as quality assessment on the candidates mined with CPD and IoU information.} 
We explain this from the view on the necessity and the role of online information separately.
1) The success of anchor-based detector demonstrates that only utilizing offline information (CPD or IoU) as metric to determine the boundary between $pos/neg$ anchors can provide a great optimization direction for the detector. However, with the number of iterations increasing, offline information  can not provide the progressive matching rules, resulting in the sub-optimal optimization direction.
Analogously, Hambox \cite{liu2019hambox} points out the same conclusion that even though adopting offline information as metric to distinguish $pos/neg$ anchros, many negative anchors have amazing regression ability. Such inconsistent phenomenon between the learnt knowledge on detector and the formulated knowledge on offline information (IoU) suggests that offline information based label assignment strategy fails to satisfy the requirements of optimization process. Therefore, combing offline and online information can provide more accurate optimization direction, due to online information can reflect the learnt knowledge on the detector. 
2) Relying too much on  the online rule to identify $pos/neg$ anchors  has an obvious drawback that it is easily over fitting on the bias assignment strategy, especially on the early stage of optimization process. Thus, rather than serving the online information as dominant criterion, we regard online rule as quality assessment tool (auxiliary role) to measure the candidates which already have mined with CPD and IoU rules. 

\begin{figure*}[t]
    \centering
     \includegraphics[width=0.9\textwidth]{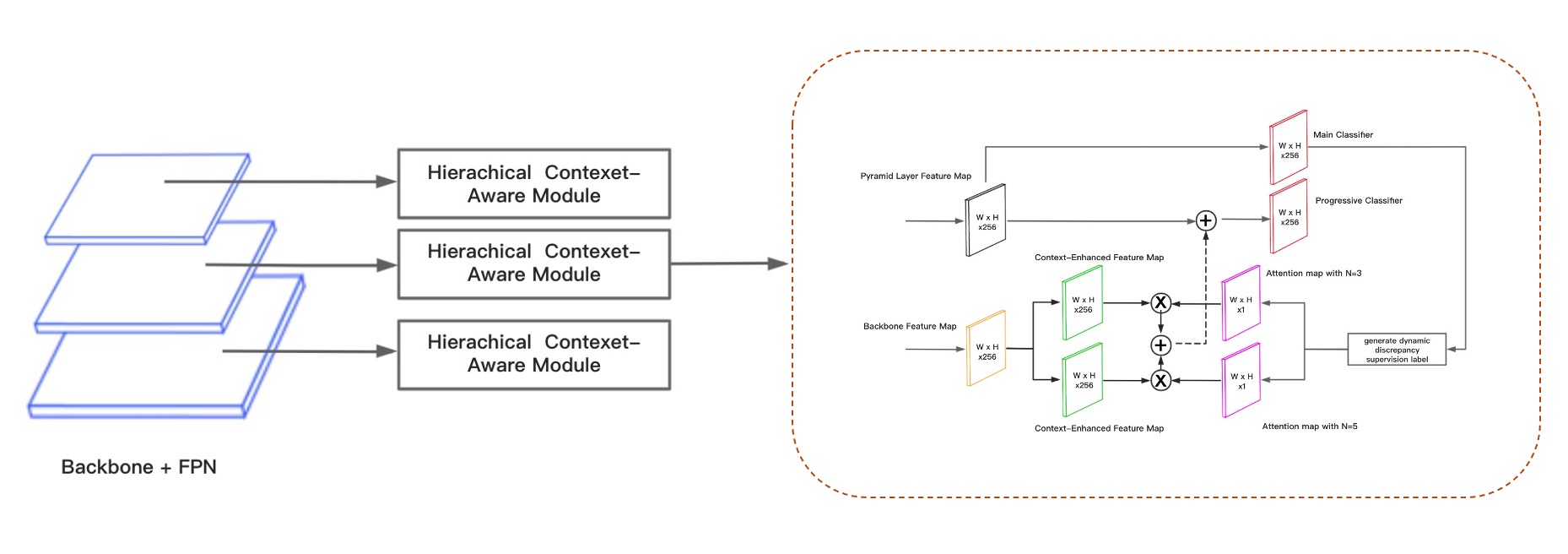}
    \caption{Hierarchical Context-Aware Module.}
    \label{img2}
\end{figure*}

\subsection{Selective Scale Enhancement Strategy}
As discussed above, previous scale-level data augmentation strategies fail to resolve the challenging problem: How to increase the detection ability on the medium and large scale faces since the training set only contains a small proportion large faces (10\%) ? To resolve this, we first analyze the relationship between the performance of each pyramid layer and the number of ground-truths it matches in the supplementary material. To our surprise, we discover an amazing conclusion:  it  is  not  accurate  that  the  more ground-truths matched in one pyramid layer, the greater  performance  of  this  pyramid  layer. Based on this instructive finding, we propose a selective scale enhancement strategy (SSE) to maximize the learning capacity of deeper pyramid layer by controlling the distribution of ground-truths among pyramid layers from $p2$ to $p7$ based on the prior statistical result.

To begin with, we introduce three prerequisites for our SSE strategy. 
(1) We firstly define the conception of the main pyramid layer and the auxiliary pyramid layer, which embraces the top-2 greatest detection abilities on the large-scale faces among all pyramid layers according to the empirical results reported in the supplementary material. As a result, p5/p6 are the main/auxiliary pyramid layer for the SSE strategy. 
(2) Then we determine the ratio of the ground-truths matched in the main and auxiliary pyramid layer. Let $r\_pi$ (i=2,3,4,5,6,7) represents the maximum performance ratio matched in the pyramid layer $pi$. The maximum performance ratio refers to that this ratio together with our proposed scale control strategy \footnote{Details shown in the supplementary materials} achieves the best performance  among all candidate ratios, that is shown on the table \ref{table_0} with bold annotation. Thus, $r\_p5$ equals to 20\%  and $r\_p6$ equals to 20\%.  
For the main pyramid layer, we define the ratio of the total ground-truths matched in the main pyramid layer as $tr\_mpl$ that equals to $r\_p5$. Moreover, we define the  ratio of the total ground-truths matched in the auxiliary pyramid layer as $tr\_apl$ that equals to (1 - $tr\_mpl$) $*$ $r\_p6$. This assignment strategy on the scale information of ground-truths can guarantee the learning capacity of the main and auxiliary pyramid layer successively. 
(3) Finally, in the table \ref{table_4}, we compute the scale range ($sr\_pi$, i=2,3,4,5,6,7) of the faces that matched in different pyramid layers. 
Note that the overlap scale range between neighbour pyramid layers is divided uniformly for convenience. 

\begin{table}[h]
\small
\renewcommand\arraystretch{1.1}
	\begin{center}
	\setlength{\tabcolsep}{10pt}
	\begin{tabular}{c|cccc}
		\hline
		   & 20\%  & 40\% & 60\% & 80\% \\
		\hline
        p4 (easy) &67.0 &75.3 &81.3 & \textbf{84.3} \\
        p4 (med) &75.3 &82.6 &\textbf{84.5} & 83.6 \\
        p5 (easy) &81.8 &\textbf{82.2} &77.9 & 73.2 \\
        p5 (med) &\textbf{85.4} &85.2 &85.0 & 83.7 \\
        p6 (easy) &\textbf{86.2} &83.1 &84.8 & 83.3 \\
        p6 (med) & \textbf{85.2} & 80.5 & 82.2 & 81.2 \\
		\hline				
	\end{tabular}
	\end{center}
\vspace{-10pt}
\caption{The results of scale control strategy on the Wider Face validation subsets.
}
\label{table_0}
\end{table}

\begin{table}[h]
\small
\renewcommand\arraystretch{1.1}
	\begin{center}
	\setlength{\tabcolsep}{20pt}
	\begin{tabular}{c|cccc}
		\hline
		   & start\underline{ }scale & end\underline{ }scale \\
		\hline
	    $sr\_p2$ & 8.4 & 20.7 \\
	    $sr\_p3$ & 20.7 & 48.2 \\
	    $sr\_p4$ & 48.2 & 106.2 \\
	    $sr\_p5$ & 106.2 & 212.4 \\
	    $sr\_p6$ & 212.4 & 420.8 \\
	    $sr\_p7$ & 420.8 & 640 \\
		\hline				
	\end{tabular}
	\end{center}
\vspace{-10pt}
\caption{The scale range of the ground-truths that matched in different pyramid layers. 
}
\label{table_4}
\end{table}

Based on 3 above prerequisites, algorithm \ref{algorithm} shows the pipeline of our SSE strategy on each training image. 1) Resize image by reshaping  the  short  side  of  image into a scale selected from scale range [640, 1280].  2) Randomly sample a face from the resized image and compute its scale $fs$. 3) Identify the target pyramid layer ($tpl$) . Randomly sample a floating-point number ($rn$)  from 0 to 1. If the value of the $rn$ is less than $tr\_p5$, we define the target pyramid layer  as $p5$. If $rn$ is over $r\_p5$ and less than $r\_p5$ + $r\_p6$,  the target pyramid layer ($tpl$) equals to p6. If  $rn$ is over  $r\_p5$ + $r\_p6$,  the target pyramid layer equals to
the random one of the pyramid layers except for $p5$ and $p6$. 4) Random select a scale from $sr\_tpl$ and compute the target resize ratio ($trr$) by $fs$ / this scale. 
5) Resize image with $trr$. 6) Define the resolution of the input image as N $\times$ N. If the resolution of the resized image is over N $\times$ N, we crop N $\times$ N area randomly as the input image and pad zero pixel if it is less than N $\times$ N.
The motivation behind our SSE strategy is as follows.
\begin{algorithm}[t!]
\small
\caption{Selective Scale Enhancement Strategy (SSE)} 
\label{algorithm}
\begin{algorithmic}[1]
\REQUIRE ~~\\
$\mathcal{I}$ is a set of all training data \\
$tr\_p5$ is a hyperparameter that represents the ratio of the total ground-truths matched in the main pyramid layer (p5). \\
$tr\_p6$ is a hyperparameter that represents the ratio of the total ground-truths matched in the auxiliary pyramid layer (p6). \\
$sr\_tpl$ is the scale range of the ground-truths that matched in the $tpl$. \\
$\mathcal{N}$ is the side of the resolution for the input image.
\ENSURE ~~\\
$\mathcal{S}$ is a set of training data augmented by our SSE strategy \\
\vspace{2mm}
\FOR{each image $i \in \mathcal{I}$}
\STATE $\mathcal{R}_{i} \leftarrow{}$ Resize image by reshaping the short side of image into a scale selected from scale range [640, 1280] randomly.\\
\STATE $\mathcal{F}_{s} \leftarrow{}$Compute the face scale that is selected from  $\mathcal{R}_{i}$ randomly. \\
\STATE $Random\_float = Random.random(0,1)$ \\
\IF{ $Random\_float  < tr\_p5$}
\STATE $tpl=p5$ \\
\ELSIF{$Random\_float <= ( tr\_p5 + tr\_p6)$} 
\STATE $tpl=p6$ \\
\ELSE
\STATE $tpl = random([p2, p3, p4, p7])$ \\
\ENDIF
\STATE compute target resize ratio: $trr = random(sr\_tpl) / \mathcal{F}_{s}$;\\
\STATE  $\mathcal{R}^{trr}_{i} \leftarrow{}$ Resize $\mathcal{R}_{i}$ with the shrink ratio $trr$; \\
\IF{ $resolution(\mathcal{R}^{trr}_{i}) > (\mathcal{N}, \mathcal{N})$}
\STATE $\mathcal{R}^{trr}_{i} \leftarrow{}$ crop $\mathcal{N} \times \mathcal{N}$ area randomly from $\mathcal{R}^{trr}_{i}$;  \\
\ELSE
\STATE $\mathcal{R}^{trr}_{i} \leftarrow{}$ expand $\mathcal{R}^{trr}_{i}$ into ($\mathcal{N}, \mathcal{N}$) by padding zero pixel;   \\
\ENDIF
\IF{$\mathcal{S}$ is None}
\STATE $\mathcal{S} \leftarrow{} \varnothing $ \\
\ENDIF
\STATE $\mathcal{S} = \mathcal{S} \cup  \mathcal{R}^{trr}_{i}$; \\
\ENDFOR
\RETURN $\mathcal{S}$;
\end{algorithmic}
\end{algorithm}

\noindent\textbf{Depending on the main and auxiliary pyramid layer to control the scale distribution of  ground-truths.} The previous designation on controlling the scale distribution of the ground-truths can be divided into following two types. MST and RCS emphasize on the uniform sampling while DAS points out an alternative view that the more small-scale data can boost the detection ability on small faces.
However, such heuristic designation can not utilize the scale information effectively. On the one hand, in the supplementary material, we find RSC and DAS achieve almost consistent performance on the Wider Face hard subset although DAS brings more small faces. On the other hand, as described in the table \ref{table_0}, we find it is not accurate that  the more ground-truths that is matched in one pyramid layer, the greater performance of this pyramid layer. These two findings demonstrate that heuristically controlling the scale distribution of ground-truths fail to satisfy the ground-truths requirements on the related pyramid layer.
Therefore, our SSE, consecutively satisfying the requirement of main and auxiliary pyramid layer on the ground-truths, is a better strategy than other heuristic designation.


\begin{figure*}[t]
    \centering
    \subfigure[AFW]{
    \includegraphics[height=0.2\textwidth]{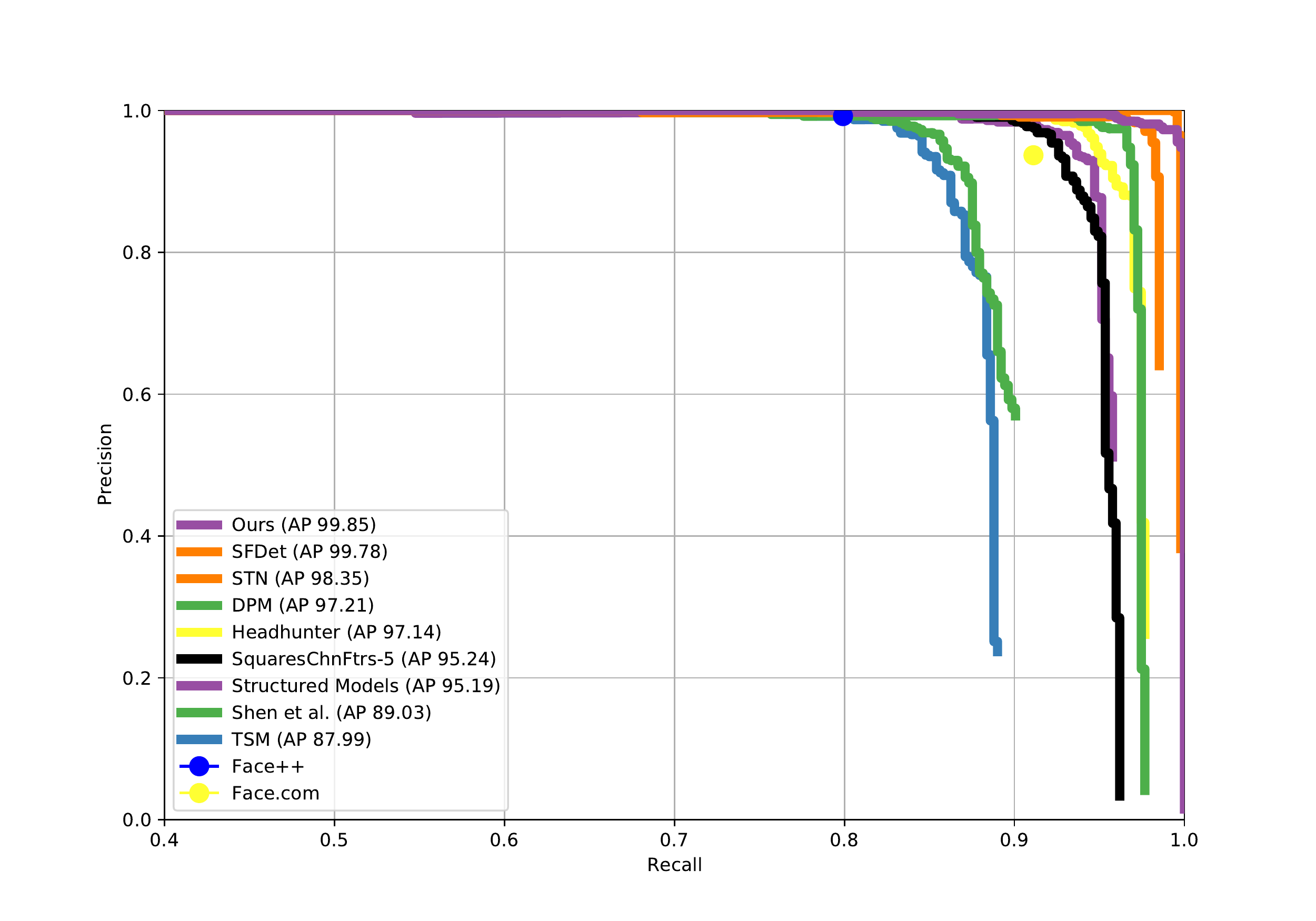}
    \label{img4_a}
    }
    \subfigure[PASCAL Face]{
    \includegraphics[height=0.2\textwidth]{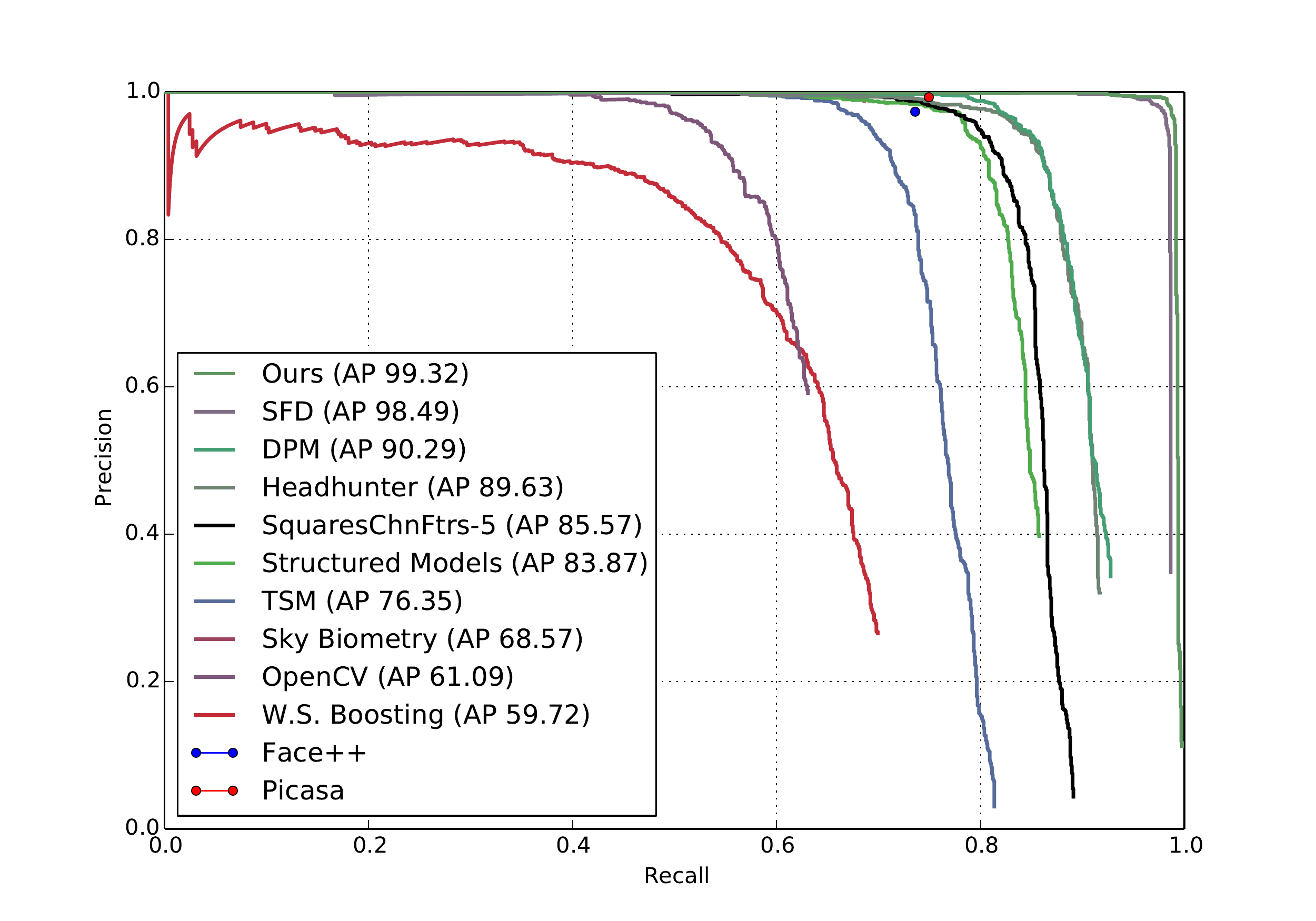}
    \label{img4_b}
    }
    \subfigure[FDDB]{
    \includegraphics[height=0.2\textwidth]{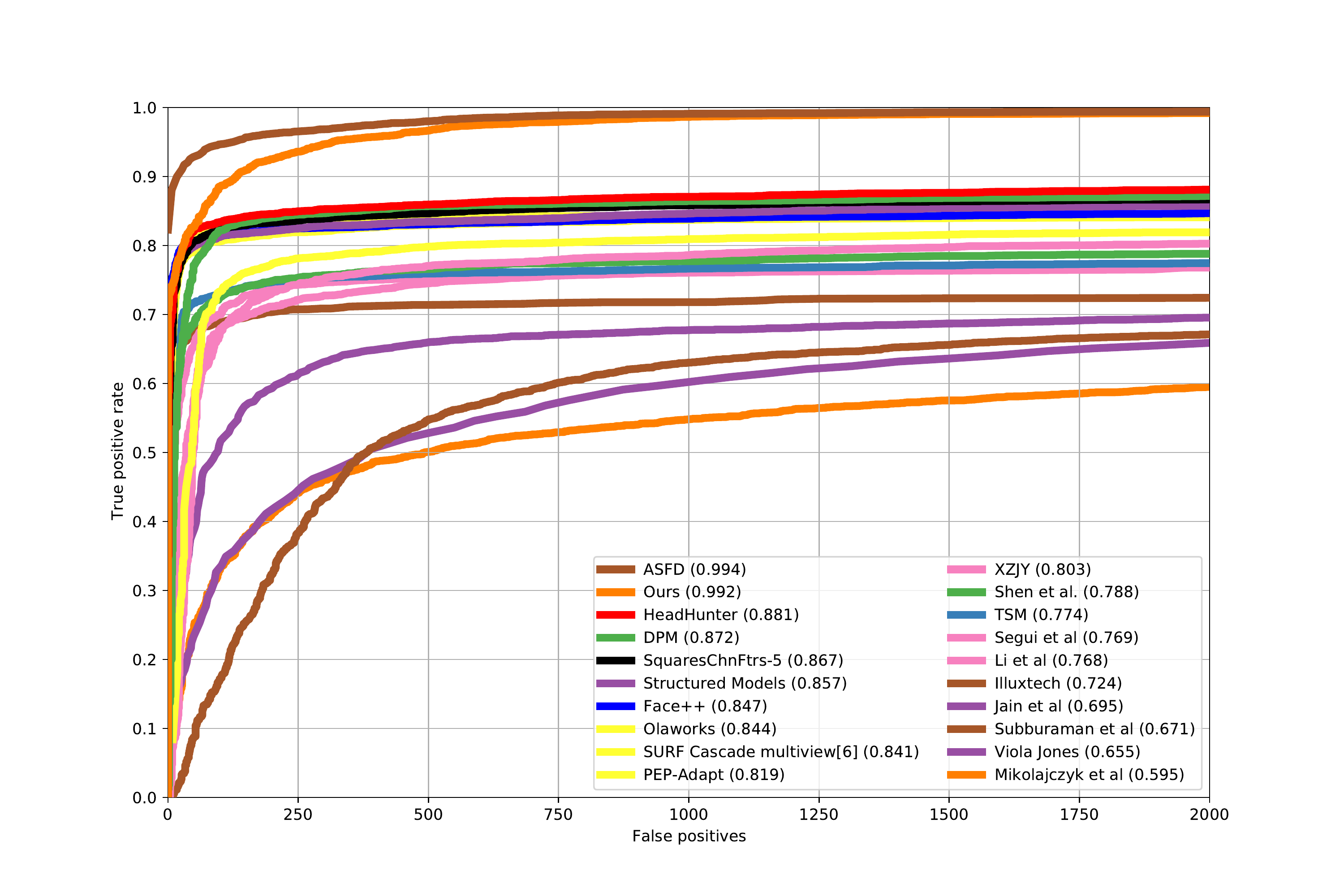}
    \label{img4_c}
    }
    \caption{Evaluation on common face detection benchmarks.}
    \label{img_4}
    \vspace{-5mm}
\end{figure*}

\subsection{Hierarchical Context-Aware Module}
As analyzed above, we find the appropriate neighbour context information is conducive to eliminating false alarms. Motivated by this finding, we propose a Hierarchical Context-Aware Module (HCAM) to distinguish false alarms away from the face by encoding the neighbour context information into related  high-confidence  anchors in Fig.~\ref{img2}. Under the rescue of neighbour context 
information, false predicted negative anchors and correct predicted positive anchors within the high-confidence anchors can be separated explicitly. The training pipeline is as follows:

1) We firstly send a batch of images into the detector and get the classification score of each anchor on the main classifier. Then we mask the position of all high-confidence anchors to generate two attention feature maps. For each attention feature map,  we assign the position of high-confidence  anchors and their related neighbour information \footnote{neighbour information takes the target anchor point as the center and contains the surrounding N$\times$N area, N is an integer greater than 0} as 1, otherwise is 0. The N are set 3 and 5 for generating different neighbour information.

2) We compute the two neighbour context information by dot-multiplying the context-enhanced backbone feature maps and  two attention feature maps, respectively. This step aims to explicitly generate more abundant context information related to the high-confidence anchors. Note that more ablative experiments and architectures on context-enhanced modules are discussed in the supplementary material.

3) We further encode two neighbour context information  into  pyramid feature map by element-wise summation operator. 

4) Finally, this combined feature map feeds into   progressive classifier to further distinguish false predicted negative anchors (false alarms)  away from correct predicted positive anchors according to Equ. \ref{equ_5}. 

\begin{equation}
    L = f_{fl}^{m}(f_c, y) + \gamma * f_{fl}^{p}(f_{c}^{com}, y_{hc}) \label{equ_5}
\end{equation}

where $f_{fl}^{m}$ and $f_{fl}^{p}$  represent the sigmoid focal loss \cite{lin2017focal} over two classes, that are applied on the main and progressive classifier, respectively. $y$ is the the label of anchor which is assigned by anchor matching strategy \cite{ren2015faster}. The weight $\gamma$ aims to balance the loss between main and progressive classifiers (we find $\gamma$=1 has a best performance). $fc$ and $f_{c}^{com}$ represent the output of main classifier and progressive classifier, separately.
$y_{hc}$ is dynamic discrepancy supervision label, that is  defined by following two steps: 1) In each iteration of the training stage, we  firstly  mask  the  position  of correctly predicted positive anchors \footnote{refer to the positive anchors which have high confidence classification score} and false predicted negative anchors \footnote{refer to the negative anchors which have high confidence classification score} according to the predicted  classification score on the main classifier at the end of forward propagation. 2) Assigning the position of  correct predicted positive anchors as positive samples, the position of false predicted negative anchors as  negative samples, the position of remaining anchors as ignore samples.

    
\begin{figure*}[t]
    \centering
    \hspace{0.3in}
    \subfigure[Val: Easy]{
    \includegraphics[width=2in]{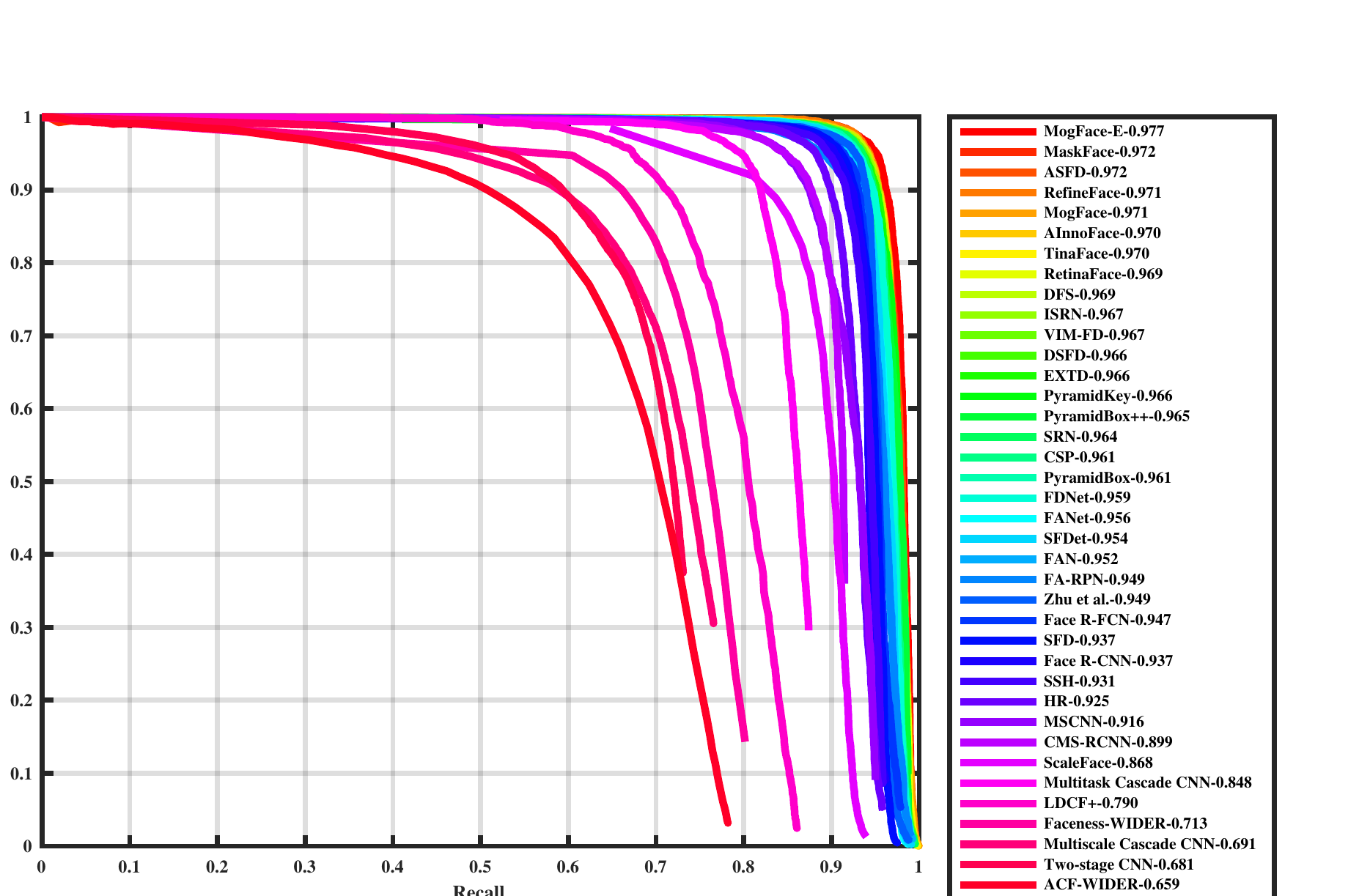}
    }
    \subfigure[Val: Medium]{
    \includegraphics[width=2in]{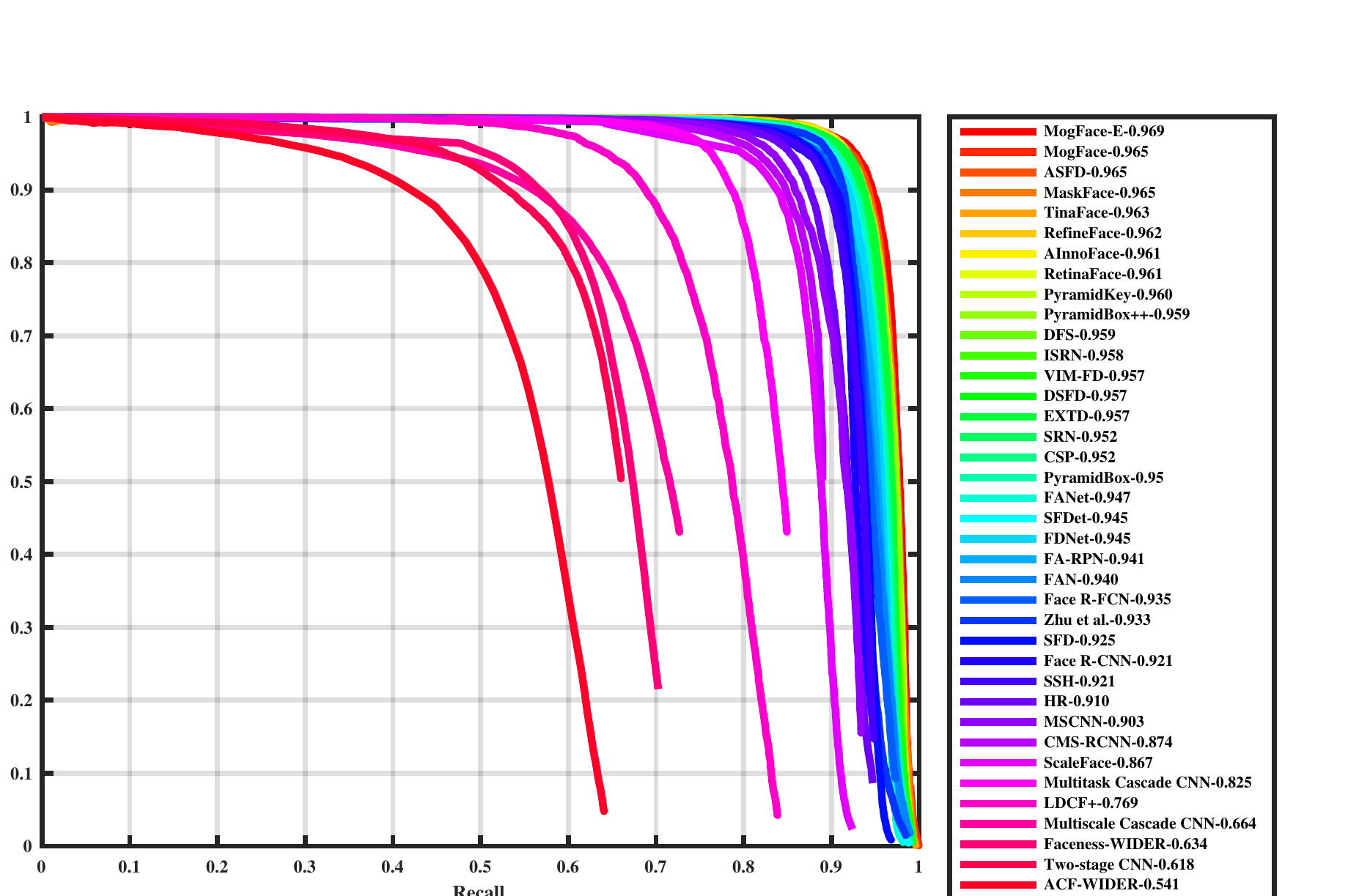}
    }
    \subfigure[Val: Hard]{
    \includegraphics[width=2in]{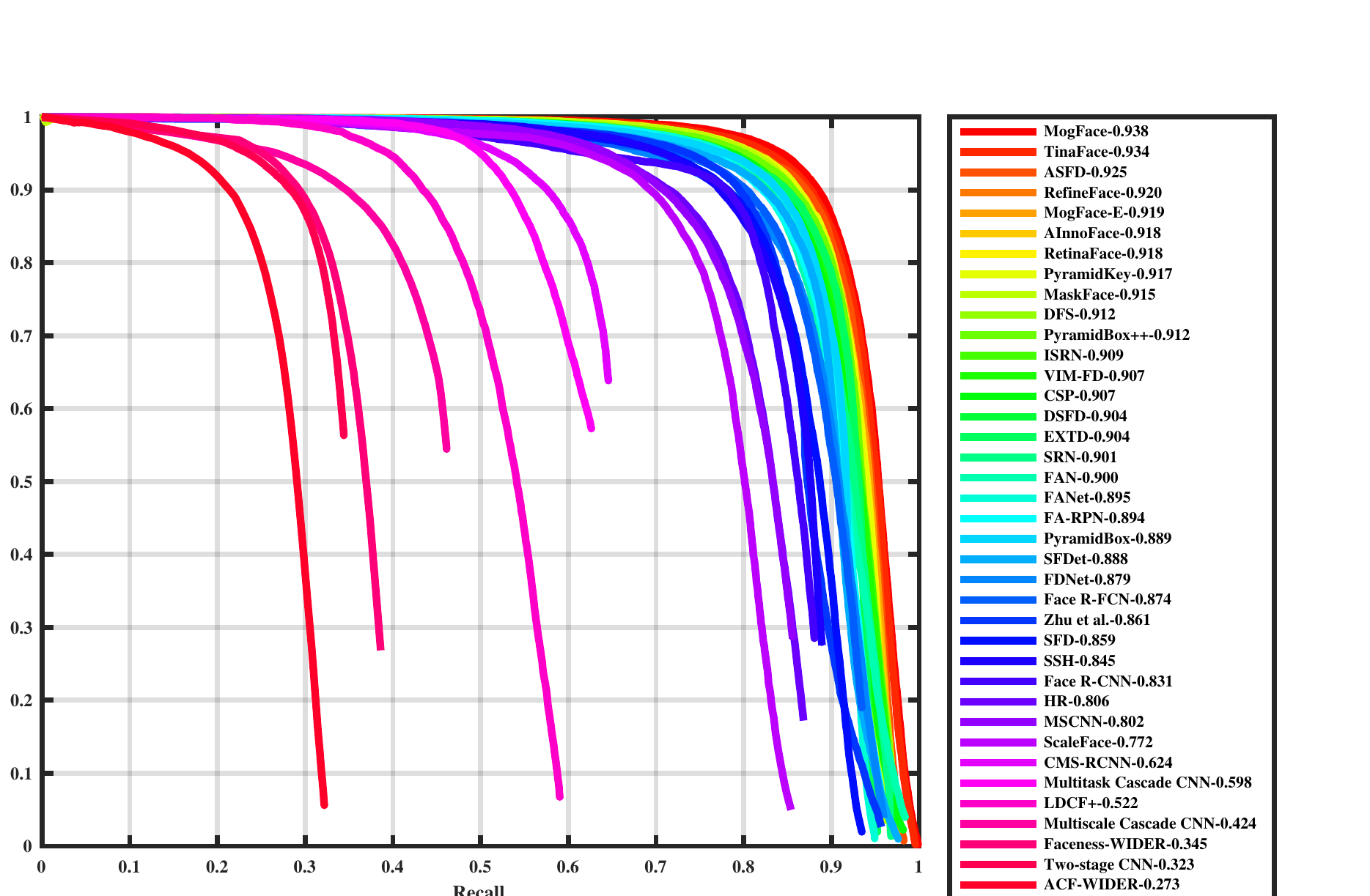}
    }
    
    \hspace{0.3in}
    \subfigure[Test: Easy]{
    \includegraphics[width=2in]{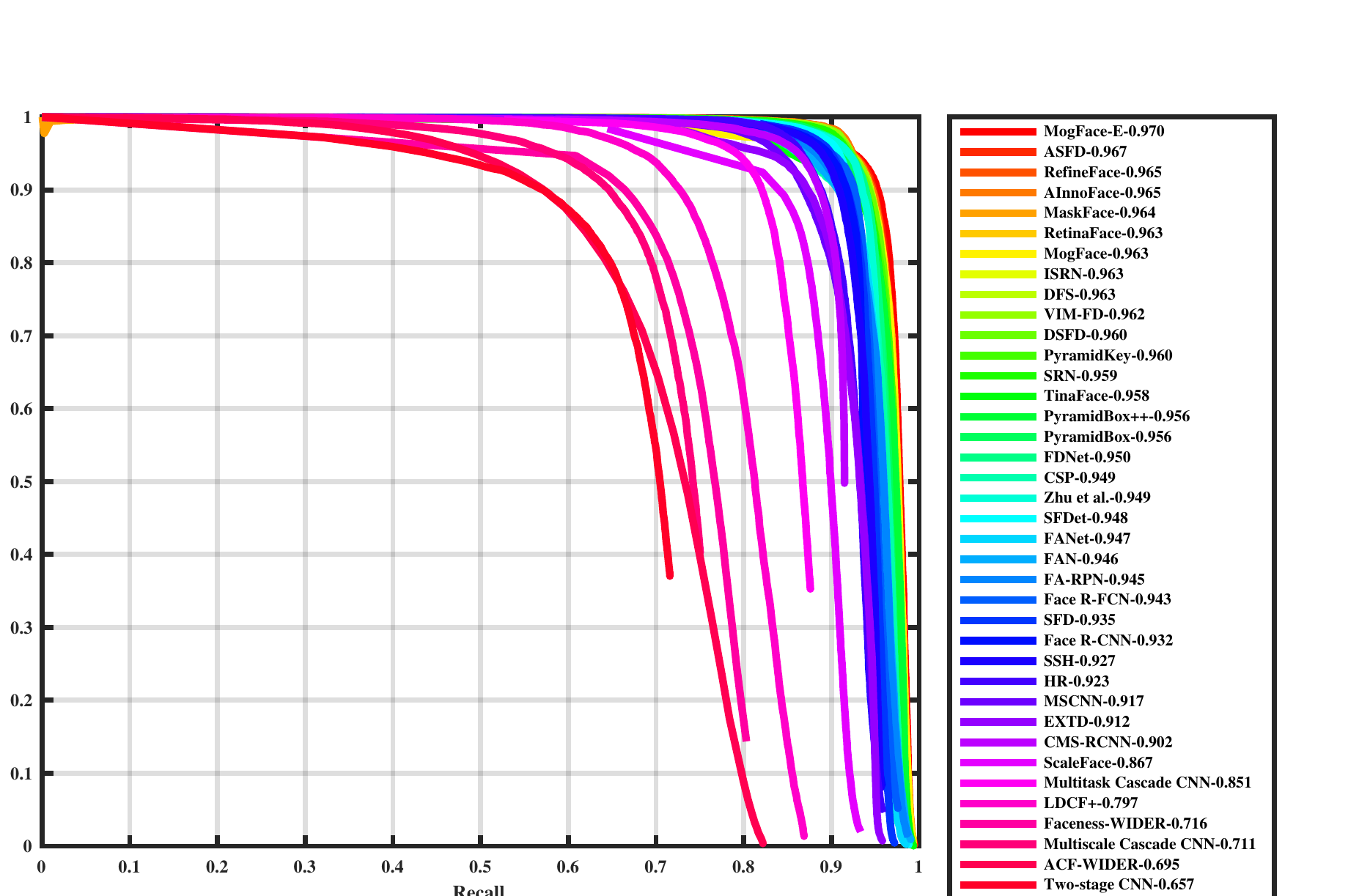}
    }
    \subfigure[Test: Medium]{
    \includegraphics[width=2in]{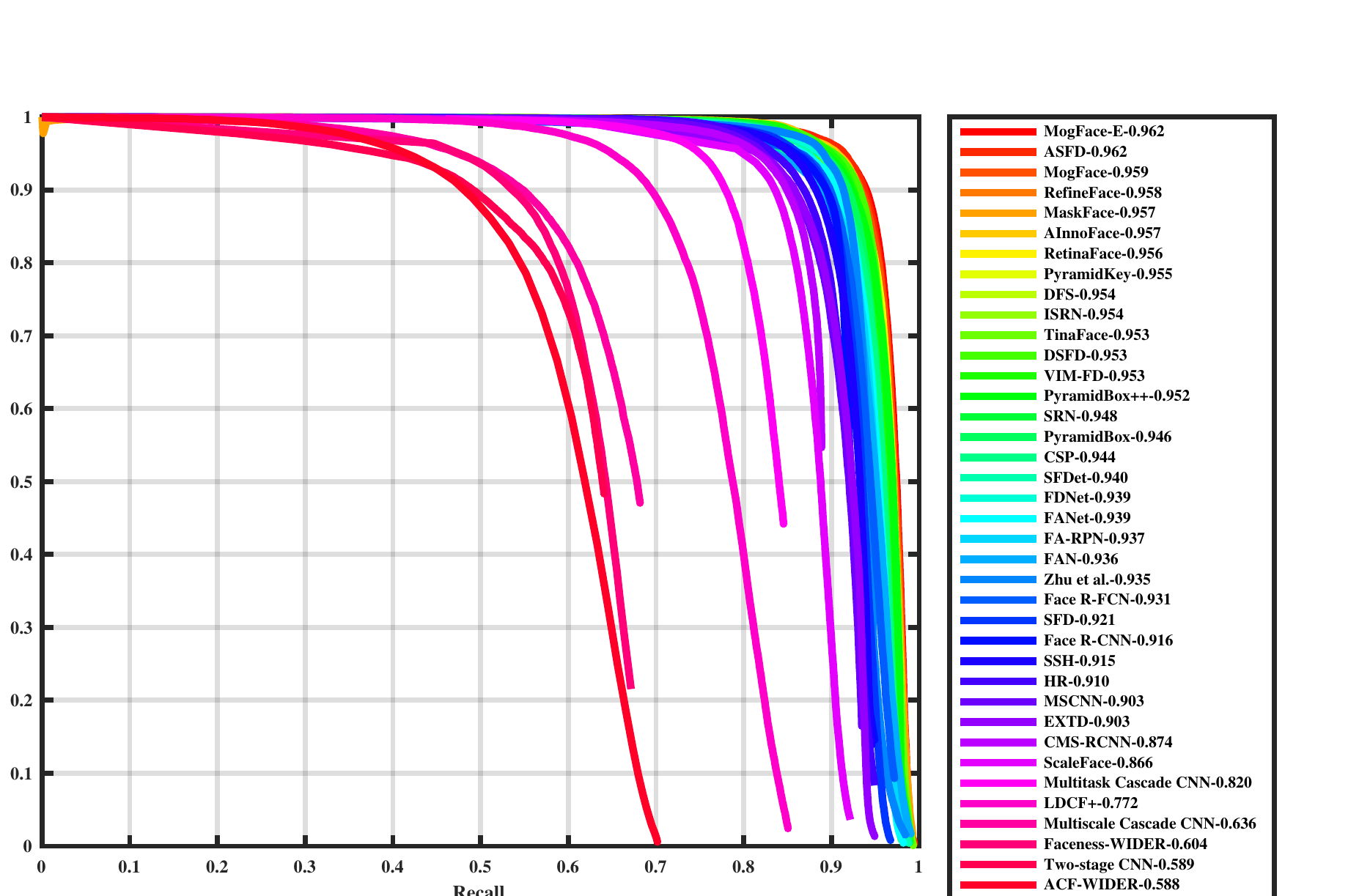}
    }
    \subfigure[Test: Hard]{
    \includegraphics[width=2in]{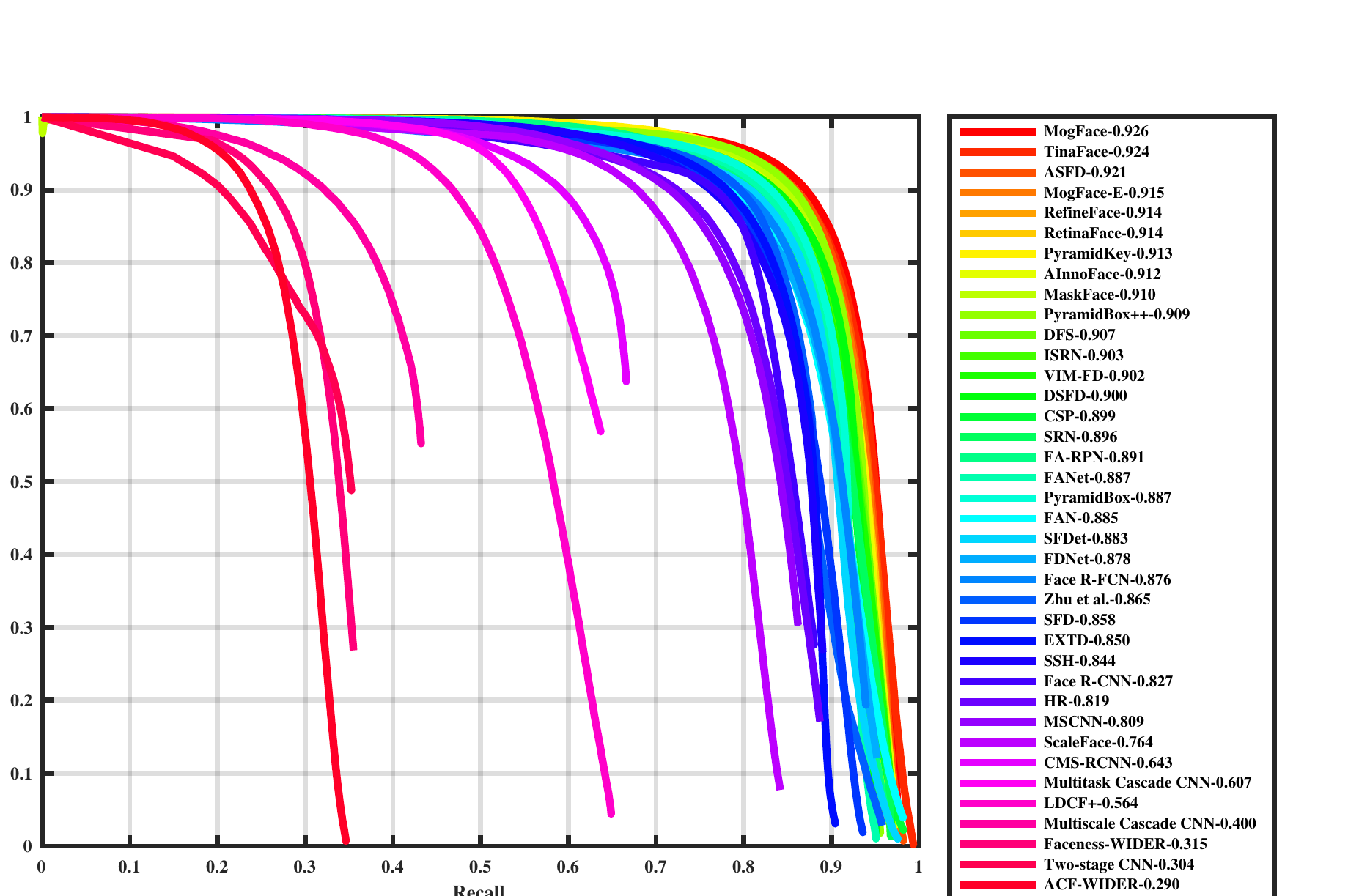}
    }
    \caption{Precision-Recall (PR) curves on Wider Face validation and testing subsets.}
    \label{img_5}
    \vspace{-2mm}
\end{figure*}

\section{Experiment}
\label{sec:experiment}
In this section, we first elaborate the implementation details of our baseline. Then, we conduct ablative experiments on the most authoritative face detection benchmark, Wider Face. Finally, we test our method on all popular face datasets, including AFW, PASCAL Face, FDDB and Wider Face. 

\subsection{Baseline Setup}
\noindent\textbf{Training Detail.} 
 We adopt SFD \cite{zhang2017s3fd} architecture with Resnet50 \cite{he2016deep} backbone as our baseline. For anchor settings,  we set 6 anchors whose scales are from the set \{16, 32, 64, 128, 256, 512\}, and all anchors’ aspect ratios are set to 1:1. The losses applied on our baseline are focal loss and smooth L1 loss with the weight ratio 1:2. 
 For optimization details, each training iteration contains 7 images per GPU on 4 NVIDIA Tesla V100s. Models are optimized by synchronized SGD. The momentum and weight decay are set to 0.9 and 5$\times$10-5, respectively. For learning rate schedule, the initial learning rate is set to 4e-3 and decreases to 4e-4 in 50000 iterations. The total iteration is 70000. 
 
 \noindent\textbf{Inference Detail.}
In the phase of inference, we adopt single scale test strategy to evaluate the experiment result. Firstly, we feed the image with original scale into the detector and then get top-5000 highest confidence bounding boxes. Then, the Non-maximum Suppression is applied with the IoU threshold 0.6 to get top 750 confident detection scores and related bounding boxes. 
\subsection{Ablation Study}
\noindent\textbf{The Effectiveness of Ali-AMS.} In the table \ref{table_5}, we 
compare our Ali-AMS with recent proposed label assignment strategies, including HAMBox \cite{liu2019hambox}, OTA \cite{ge2021ota} and ATSS \cite{zhang2020bridging}.  Our Ali-AMS achieves the best performance among all recent label assignment strategies on the Wider Face hard subset with outperforming baseline, HAMBox, ATSS, OTA by 0.8\%, 0.4\%, 1.0\%, 2.6\% AP score, respectively.

\begin{table}[h]
\small
\renewcommand\arraystretch{1.1}
	\begin{center}
	\setlength{\tabcolsep}{10pt}
	\begin{tabular}{c|ccc}
		\hline
		Method   & $\text{Easy}$  & $\text{Medium}$ & $\text{Hard}$ \\
		\hline
        Baseline  &94.6  &93.4 & 86.5 \\
        Baseline + HAMBox  &94.5  &\textbf{93.8} & 86.9 \\
        Baseline + ATSS &94.5	&93.2	&86.3 \\
        Baseline + OTA &93.3  &91.7 & 84.7 \\
        Baseline + Ali-AMS  &\textbf{94.6}  &93.6 & \textbf{87.3}\\   
		\hline				
	\end{tabular}
	\end{center}
\vspace{-10pt}
\caption{Results of Ali-AMS on the Wider Face validation subsets.
}
\label{table_5}
\end{table}
\noindent\textbf{The Effectiveness of SSE.} In the table \ref{table_6}, we compare our SSE with other scale-level data augmentation strategies, including data-anchor-sampling, random square crop and multi scale training strategy. Our SSE outperforms others by 1.0\%, 0.7\% AP on the Wider Face validation easy, medium subset respectively. Such tremendous enhancement on detecting large-scale faces help our MogFace-E achieve 4 champions on the Wider Face official leader-board.
\begin{table}[h]
\small
\renewcommand\arraystretch{1.1}
	\begin{center}
	\setlength{\tabcolsep}{10pt}
	\begin{tabular}{c|ccc}
		\hline
		Method   & $\text{Easy}$  & $\text{Medium}$ & $\text{Hard}$ \\
		\hline
        Baseline (w/o DAS)  &92.2	& 90.5 & 81.4 \\
        Baseline + MST &93.3	& 91.5	&83.6 \\
        Baseline + DAS &94.6  &93.4 & \textbf{86.5} \\
        Baseline + SSE  &\textbf{95.6}  &\textbf{94.1} & - \\   
		\hline				
	\end{tabular}
	\end{center}
\vspace{-10pt}
\caption{Results of SSE on the Wider Face validation subsets.
}
\label{table_6}
\end{table}

\noindent\textbf{The Effectiveness of HCAM.} Table \ref{table_8} presents the performance and the number of false alarms (NFA) on our Hierarchical Context-Aware false alarms  Module with different N (defined in the footnote 5). The Hierarchical Context-Aware Module achieves the best performance (87.4 \% AP on the Wider Face Validation hard subset) and contains the minimal false alarms, when N=3 and 5. Note that the  context-aware module can also only contain one neighbour context information, e.g. N=3 or N=5.

\begin{table}[h]
\small
\renewcommand\arraystretch{1.1}
	\begin{center}
	\setlength{\tabcolsep}{6pt}
	\begin{tabular}{c|ccccc}
		\hline
		Method   & N  &$\text{Easy}$  & $\text{Medium}$ & $\text{Hard}$ &$\text{NFA}$ \\
		\hline
		Baseline  & - &94.6  &93.4 &86.5 & 948 \\
		HCAM    &3  & 94.9 &94.0 & 86.8 & 532\\
		HCAM    &5  & 94.8 &94.1 & 87.0 & 476\\
		HCAM    &3 and 5 &\textbf{95.1}	& \textbf{94.2}	& \textbf{87.4}  &\textbf{192}\\
		\hline				
	\end{tabular}
	\end{center}
\vspace{-10pt}
\caption{Results of our Hierarchical Context-Aware Module on the Wider Face validation subsets.
}
\label{table_8}
\end{table}

\subsection{Evaluation on Common Benchmarks}

In this subsection, we  compare our MogFace with state-of-the-arts methods on the common face detection benchmarks, including AFW \cite{zhu2012face}, Pascal Face \cite{yan2014face}, FDDB \cite{jain2010fddb} and Wider Face \cite{yang2016wider}. We train the MogFace-E (Ali-AMS, HCAM, SSE) and MogFace (Ali-AMS, HCAM) with some excellent modules introduced by Hambox \cite{liu2019hambox} detector on the  Wider Face training dataset, including SSH head \cite{najibi2017ssh}, Pyramid Anchor \cite{tang2018pyramidbox} and deep head \cite{lin2017focal}. 

\noindent \textbf{AFW Dataset.} This dataset contains 205 images with 473 annotated faces. As shown in Fig. \ref{img4_a},  our Mogface significantly outperforms other methods by 1.0\% AP at least.

\noindent \textbf{PASCAL Face Dataset.} This dataset contains 851 images with 1335 annotated faces. Fig.~\ref{img4_b} shows that our method achieves the state-of-the-art results by outperforming the second one with $0.8\%$ AP. 

\noindent \textbf{FDDB Dataset.} This dataset has 2,845 images with 5,171 annotated faces. Most of them have low image resolutions and complicated scenes
. Fig.~\ref{img4_c} shows that our MogFace achieves the highest (99.2\%) performance.

\noindent\textbf{Wider Face Dataset.} We test our MogFace and MogFace-E with Multi-scale results ensemble strategy on the Wider Face validation and test set. As shown in Fig. \ref{img_5}, we plot a precision-recall curve according to the official tool on the Wider Face validation set.  As for Wider Face test set, we submit the detection bounding boxes and corresponding scores to the official server to get the precision-recall curves that is shown in Fig. \ref{img_5}. Our method achieves 97.7\% (Easy), 96.9\% (Medium), 93.8\% (Hard) AP performance on the Wider Face validation set and 97.0 \% (Easy), 96.2\% (Medium), 92.6\% (Hard) AP performance on the Wider Face test set. Comparing with other sota approaches  \cite{hu2017finding,wang2017detecting,li2019dsfd, wang2017face, chi2019selective,zhang2018face, zhu2018seeing, cai2016unified,zhu2017cms,zhang2020asfd}, our method outperforms them by  0.5\% (Validation Easy), 0.5 \% (Validation Medium), 0.4 \% (Validation Hard), 0.3 \% (Test Easy), 0.2 \% (Test Hard). Such tremendous enhancement in all scenarios demonstrates the superiority of our MogFace.

\section{Conclusion}
In this paper,  we first point out that the success experience from generic object detection fails to provide effective solutions on  label assignment, scale-level data augmentation, and eliminating face alarms in the field of face detection. 
Thereby, in order to advance the development of face detectors, we resolve three aforementioned challenges by proposing Adaptive Online Incremental Anchor Mining Strategy, Selective Scale Enhancement Strategy, Hierarchical Context-Aware Module, respectively. Finally, benefiting from prominent solutions of our MogFace,  we achieve six champions on the Wider Face dataset, which continues to this day. 

\newpage

{\small
\bibliographystyle{ieee_fullname}
\bibliography{egbib}
}

\end{document}


\title{Supplementary Material}

\maketitle

\maketitle
\section{Description}
Due to the page limited, we will clarify  the quantitative and qualitative analysis on the scale-level data augmentation strategies, and  more ablative experiments and architectures on context-enhanced modules of HCAM in the supplementary material. 

\begin{figure*}[t]
    \subfigure[]{
    \includegraphics[width=0.38\textwidth]{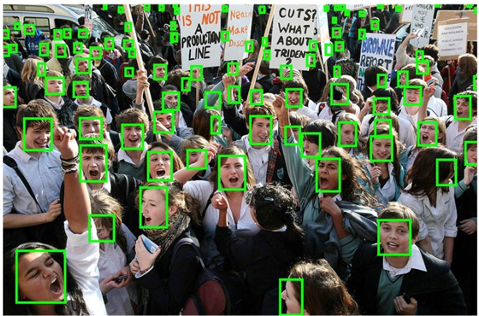}
    \label{img1_a}
    }
    \hspace{0.2in}
    \subfigure[]{
    \includegraphics[width=0.25\textwidth]{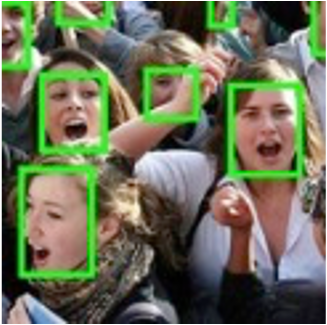}
    \label{img1_b}
    }
    \hspace{0.2in}
    \subfigure[]{
    \includegraphics[width=0.25\textwidth]{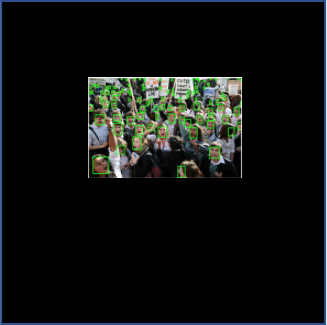}
    \label{img1_b}
    }
    \caption{An image is augmented by multi-scale training and data-anchor-sampling respectively. (a) An original image. (b) The image in (a) is augmented by the random crop strategy. The native image information is less than the original image. Note that this image may blur the image when the expand ratio is large. (c) The image in (a) is augmented by the data-anchor-sampling strategy. This brings a  large amount of padding area, that reduces the learning difficulty of the detector on negative anchors. Thus, the fore-ground information in native image information is reduced remarkably.  }
    \label{img2}
\end{figure*}

\section{Analysis on Scale-level Data Augmentation}
\label{sec:Analysis}
In this section, we firstly implement Random Square Crop (RSC), Data Anchor Sampling (DAS) and Multi Scale Training (MST) on the proposed baseline and find the former two strategies have better performance than the last one. Secondly, based on this experiment result, we further analyze the reason of why does the MST strategy perform poorly on resolving extreme scale variance challenge? Thirdly, in order to utilize the scale information of the training data,  we  analyze the relationship between the scale information and the detector performance.

 \noindent\textbf{Implementation Detail of Scale-level Data Augmentation strategies.}
\begin{itemize}
\setlength{\itemsep}{0pt}
    \item Multi-scale-training: For each image in the training data, we resize it by reshaping the short side of image into a scale selected from predefined scale range [640, 1280] randomly.
    \item Data-anchor-sampling: 1) Randomly select a face and compute its scale $fs$ 2) Choose the nearest scale with this face
    scale from the set  $\left\{16, 32, 64, 128, 256, 512\right\}$ 3) Uniformly random select the scale $sr$ from the set $\left\{16, 32, ..., nearest\_scale \right\}$ and compute the target ratio $tr$ by $sr$ / $fs$ . 4) Resize the image with this target ratio $tr$. 5) If the resolution of resized image is over 640 $\times$ 640, we crop 640 $\times$ 640 area randomly as the input image and pad zero pixel if it is less than 640 $\times$ 640.
    \item Random Square Crop: Random crop a square area from the image with the scale that equals to multiplying the short side scale and a factor selected from $\left\{0.1, 0.3, 0.5, 0.7, 0.9\right\}$ randomly. 
\end{itemize}
 
\subsection{Difference among RSP, DAS and MST.}
As shown in Table \ref{table_1}, we display the performance of RSP, DAS and MST on our baseline detector, where  RSP and DAS achieve almost consistent performance while the MST only achieves 83.60\% AP on the Wider Face validation hard subset. Considering the tremendous performance gap , we summary the two differences among them: (1) MST brings more abundant scale information than RSP and DAS. (2) As shown in 2. \ref{img2}, RSP and DAS both reduce the native image information on the each image of training data by cropping the local square patch  from  the  original  image  patch  and  bringing  a  large amount of padding area separately. These two differences can be further interpreted as that comparing with RSP and DAS, MST introduces more scale information and native image information. 



\begin{table}[h]
\small
\renewcommand\arraystretch{1.1}
	\begin{center}
	\setlength{\tabcolsep}{10pt}
	\begin{tabular}{c|ccc}
		\hline
		Method   & $\text{Easy}$  & $\text{Medium}$ & $\text{Hard}$ \\
		\hline
        Baseline   &92.2	& 90.5 & 81.4 \\
        Baseline + MST &93.3 & 91.5	&83.6 \\
        Baseline + DAS &94.6  &93.4 &86.5 \\
        Baseline + RSC &94.8	& 94.1	&86.4 \\
        Baseline + MST + RSC &94.8  &94.2 &86.5\\       
		\hline				
	\end{tabular}
	\end{center}
\vspace{-10pt}
\caption{Results of scale-level data augmentation strategies on the Wider Face validation subsets.
}
\label{table_1}
\end{table}

Then, we conduct a  experiment to explore whether scale information or native image information \footnote{Native Image Information refers to the fore-ground and back-ground information of an image} causes the significant performance gap. As shown in Table \ref{table_1}, we combine MST with RSC to help the detector embrace more scale information and few native image information. Comparing with the detector only with MST, the performance increases 4.3\% AP on the Wider Face validation hard subset, that demonstrates the less native image information can provide appropriate knowledge for the detector. Simultaneously, this experiment result is almost consistent with the detector adopting RSC, which explicitly demonstrates the  scale information provided by the MST strategy is hard for the detector to absorb.  It can be concluded from the experiment results that simplex (less) native image information is conducive to the face detector facing extreme scale variance and the scale information can not be assimilated by the detector effectively.

\subsection{Analyze the Relationship between the Scale Information and the Detector Performance. }
In our perspectives, to help the detector absorb the scale information effectively, we need answer the following two questions firstly: (1) What is the relationship between the performance of each pyramid layer and the number of ground-truths it matches? (2) Can the larger shrink or expand ratio of the image provide reliable scale information?

In order to assist investigating the first question, we propose a scale control strategy based on the dichotomy that can control the ratio $r_i$ of the ground-truth matched in the target pyramid layer $p_i$. (1) Select a  middle scale $s_i$ from the interval $[start\_s, end\_s]$.  (2) For each image in the training data, random sample a ground-truth from it and get a shrink ratio $sr$ by $s_i$ / its scale. (3) Resize the image with shrink ratio $sr$. (4) Compute the ratio $r_c$ of the ground-truth  matched in the pyramid layer $p_i$ under current setting. (5) If $|r_c - r_i| < 0.05$, training the detector with scale-level data augmentation strategy like in the step (2) and (3), break; if $r_c > r_i$, $end\_s = r_c$, restart from step 1; if $r_c < r_i$, $start\_s = r_c$, restart from step 1. Based on this strategy, we further train the detector by controlling the ratio of ground-truths matched on a certain pyramid layer. For instance, when $s_i$ equals to 21, the $p_2$ can match 80\% ground-truth and the detector achieves 81.79 \% AP on the Wider Face validation hard subset. Note that the hard subset contains a large amount of small faces, so it is appropriate for the evaluation of the $p_2$,$p_3$ learning capacity. Similarly, medium (easy) subset is appropriate for $p_4$, $p_5,p_6$ ($p_5$, $p_6$). As results reported in the table \ref{table_2},  we  get a new appreciation that it is not accurate that  the more ground-truths that is matched in a single pyramid layer, the greater performance of this pyramid layer.

\begin{table}[h]
\small
\renewcommand\arraystretch{1.1}
	\begin{center}
	\setlength{\tabcolsep}{10pt}
	\begin{tabular}{c|cccc}
		\hline
		   & 20\%  & 40\% & 60\% & 80\% \\
		\hline
		p2 (hard) &81.79 &\textbf{82.17} &77.85 & 73.23 \\
        p3 (hard) &74.82 &76.52 &\textbf{77.15} & 71.94 \\
        p4 (easy) &67.01 &75.28 &81.28 & \textbf{87.31} \\
        p4 (med) &75.28 &82.60 &\textbf{84.48} & 83.62 \\
        p5 (easy) &81.79 &\textbf{82.17} &77.85 & 73.23 \\
        p5 (med) &\textbf{85.44} &85.16 &84.97 & 83.68 \\
        p6 (easy) &\textbf{86.17} &83.12 &84.78 & 83.34 \\
        p6 (med) & \textbf{85.24} & 80.47 & 82.17 & 81.22 \\
		\hline				
	\end{tabular}
	\end{center}
\vspace{-10pt}
\caption{The results of scale control strategy on the Wider Face validation subsets.
}
\label{table_2}
\end{table}

To investigate the second question, we revise the step 4 in data anchor sampling as follows: if $tr > r\_th$, $tr = r\_th$. if $tr < 1 / r\_th$, $tr = 1 / r\_th$. Thus, $r\_th$ controls the maximum shrink ratio. 
We show the results on  the detector with different $r\_th$ in table \ref{table_2}. The performance is almost consistent among different $r\_th$. Thus,  the large expand/shrink ratio of the image can also provide reliable scale information absolutely. Thus, in our SSE, we neglect to add any constraints on the maximum expand/shrink ratio of the image. 

\begin{table}[h]
\small
\renewcommand\arraystretch{1.1}
	\begin{center}
	\setlength{\tabcolsep}{10pt}
	\begin{tabular}{c|ccc}
		\hline
		$r\_th$   & Easy &Medium &Hard \\
		\hline
        2 &94.8	&93.7 & 86.5 \\
        4 &94.7 &93.6 &86.4\\
        8 &94.7 &93.4 & 86.2 \\ 
        16 &94.9 & 93.5 &86.4 \\
        32 &94.6 & 93.8 & 86.5 \\
        64 &94.7 & 93.6 & 86.4 \\
		\hline				
	\end{tabular}
	\end{center}
\vspace{-10pt}
\caption{The results of the detector with different $r\_th$ on the Wider Face validation subsets. 
}
\label{table_3}
\end{table}

\subsection{Ablative Experiments and Architectures on Context-Enhanced Modules}
As described in the step 4 of Hierachical Context-Aware Module, we introduce the Context-Enhanced Module to explicit encode context information on the backbone feature map with 3 types, atrous spatial pyramid pooling (aspp), detection head module (SSH-DH)in SSH and single 3x3 convolution layer (SCL). As shown in the table \ref{table_3}, single 3x3 convolution layer can bring 1.1 \% enhancement on the Wider Face hard subset. Comparing with another two Context-Enhanced modules, 3x3 convolution layer achieves the best trade-off between accuracy and computation cost. Thus, our HCAM adopt single 3x3 convolution layer as Context-Enhanced Module.
\begin{table}[h]
\small
\renewcommand\arraystretch{1.1}
	\begin{center}
	\setlength{\tabcolsep}{10pt}
	\begin{tabular}{c|ccc}
		\hline
		Method   & $\text{Easy}$  & $\text{Medium}$ & $\text{Hard}$ \\
		\hline
        Baseline  &94.6  &93.4 & 86.5 \\
        Baseline + ASPP &\textbf{95.5}	&\textbf{94.7}	&\textbf{87.9} \\
        Baseline + SSH-DH &95.2  &94.4 & 87.8 \\
        Baseline + SCL &95.3  &94.4 & 87.6\\   
		\hline				
	\end{tabular}
	\end{center}
\vspace{-10pt}
\caption{Results of different Context-Awared Module on the Wider Face validation subsets.
}
\label{table_3}
\end{table}

{\small
\bibliographystyle{aaai22}
\bibliography{egbib}
}